\documentclass[sigconf,10pt]{acmart}
\AtBeginDocument{%
  \providecommand\BibTeX{{%
    \normalfont B\kern-0.5em{\scshape i\kern-0.25em b}\kern-0.8em\TeX}}}

\usepackage{xcolor}
\usepackage{soul}
\usepackage{subcaption}
\usepackage{xspace}
\usepackage{tabu}
\usepackage{extarrows}
\usepackage{amsmath, nccmath}
\usepackage{geometry}
\usepackage{enumitem}

\definecolor{ilias_color_TM}{RGB}{191, 232, 255}
\definecolor{stelios_colour}{RGB}{144, 238, 144}
\definecolor{light_red}{RGB}{255, 204, 204}

\newcommand{\tool}{PersEPhonEE\xspace}

\newif\ifcomment

\commenttrue
\commentfalse

\ifcomment
\newcommand{\stelios}[1]{\sethlcolor{stelios_colour}\hl{[\textbf{Stelios:} #1]}}
\newcommand{\steve}[1]{\sethlcolor{cyan}\hl{[\textbf{Steve:} #1]}}
\newcommand{\manote}[1]{\sethlcolor{pink}\hl{[\textbf{Mario:} #1]}}
\newcommand{\il}[1]{\sethlcolor{ilias_color_TM}\hl{[\textbf{Ilias:} #1]}}
\newcommand{\cut}[1]{\sethlcolor{light_red}\hl{[#1]}}
\newcommand{\blue}[1]{\textcolor{blue}{#1}} %
\else
\newcommand{\stelios}[1]{}
\newcommand{\steve}[1]{}
\newcommand{\il}[1]{}
\newcommand{\manote}[1]{}
\newcommand{\cut}[1]{}
\newcommand{\blue}[1]{\textcolor{black}{#1}}
\fi

\usepackage{background}
\backgroundsetup{angle=0,
    scale=1,
    color=black,
    firstpage=true,
    position=current page.north,
    hshift=0pt,
    vshift=-20pt,
    contents={\ifnum\value{page}=1 PREPRINT: Accepted at the 22nd International Workshop on Mobile Computing Systems and Applications (HotMobile), 2021 \else \fi}
}

\title{It's always personal: Using Early Exits for Efficient On-Device CNN Personalisation}

\author{{Ilias Leontiadis*$^\dagger$, Stefanos Laskaridis*$^\dagger$, Stylianos I. Venieris*$^\dagger$, Nicholas D. Lane}$^{\dagger,\ddagger}$}

\affiliation{\institution{$^\dagger$Samsung AI Center, Cambridge\hspace{+0.75cm}$^\ddagger$University of Cambridge}{\Small\textit{{* Indicates equal contribution.}}}}

\email{{i.leontiadis, stefanos.l, s.venieris, nic.lane}@samsung.com}

\authorwithoutinstuition{{Ilias Leontiadis, Stefanos Laskaridis, Stylianos I. Venieris, Nicholas D. Lane}}

\newif\ifacmversion
\acmversiontrue

\ifacmversion

\copyrightyear{2021}
\acmYear{2021}
\setcopyright{acmcopyright}\acmConference[HotMobile '21]{The 22nd International Workshop on Mobile Computing Systems and Applications}{February 24--26, 2021}{Virtual, United Kingdom}
\acmBooktitle{The 22nd International Workshop on Mobile Computing Systems and Applications (HotMobile '21), February 24--26, 2021, Virtual, United Kingdom}
\acmPrice{15.00}
\acmDOI{10.1145/3446382.3448359}
\acmISBN{978-1-4503-8323-3/21/02}

\usepackage{natbib}

\else

\usepackage{background}
\backgroundsetup{angle=0,
    scale=1,
    color=black,
    firstpage=true,
    position=current page.north,
    hshift=0pt,
    vshift=-20pt,
    contents={\ifnum\value{page}=1 PREPRINT: Accepted at the 26th Annual International Conference on Mobile Computing and Networking (MobiCom), 2020 \else \fi}
}
\fi

\ifacmversion
\else
\renewcommand\footnotetextcopyrightpermission[1]{} % removes footnote with conference information in first column
\setcopyright{none}
\fi

\begin{abstract}

\vspace{-0.25em}
On-device machine learning is becoming a reality thanks to the availability of powerful  hardware %(\textit{e.g.}~ mobile GPUs and NPUs) 
and model compression techniques. Typically, these models are pretrained on large GPU clusters and have enough parameters to generalise across a wide variety of inputs.
In this work, we observe that a much smaller, \emph{personalised} model can be employed to fit a specific scenario, resulting in both higher accuracy and faster execution. 
Nevertheless, on-device training is extremely challenging, imposing excessive computational and memory requirements even for flagship smartphones. At the same time, on-device data availability might be limited and samples are most frequently unlabelled. 

To this end, we introduce \tool, a framework that attaches early exits on the model and personalises them on-device. These allow the model to progressively bypass a larger part of the computation as more personalised data become available. Moreover, we introduce an efficient on-device algorithm that trains the early exits in a semi-supervised manner at a fraction of the whole network's personalisation time.
Results show that \tool boosts accuracy by up to 15.9\% while dropping the training cost by up to $2.2\times$ 
and inference latency by $2.2$-$3.2\times$ on average for the same accuracy, depending on the availability of labels on-device.

\vspace{-0.2em}

\end{abstract}

\begin{document}

\fancyhead{}
\maketitle

\vspace{-0.3cm}
\section{Introduction}
\label{sec:intro}
\vspace{-0.1cm}

\begin{figure}[t]
      \centering
      \vspace{-0.1cm}
      \includegraphics[trim={8.7cm 4.2cm 8.7cm 4.5cm},clip,width=0.85\columnwidth]{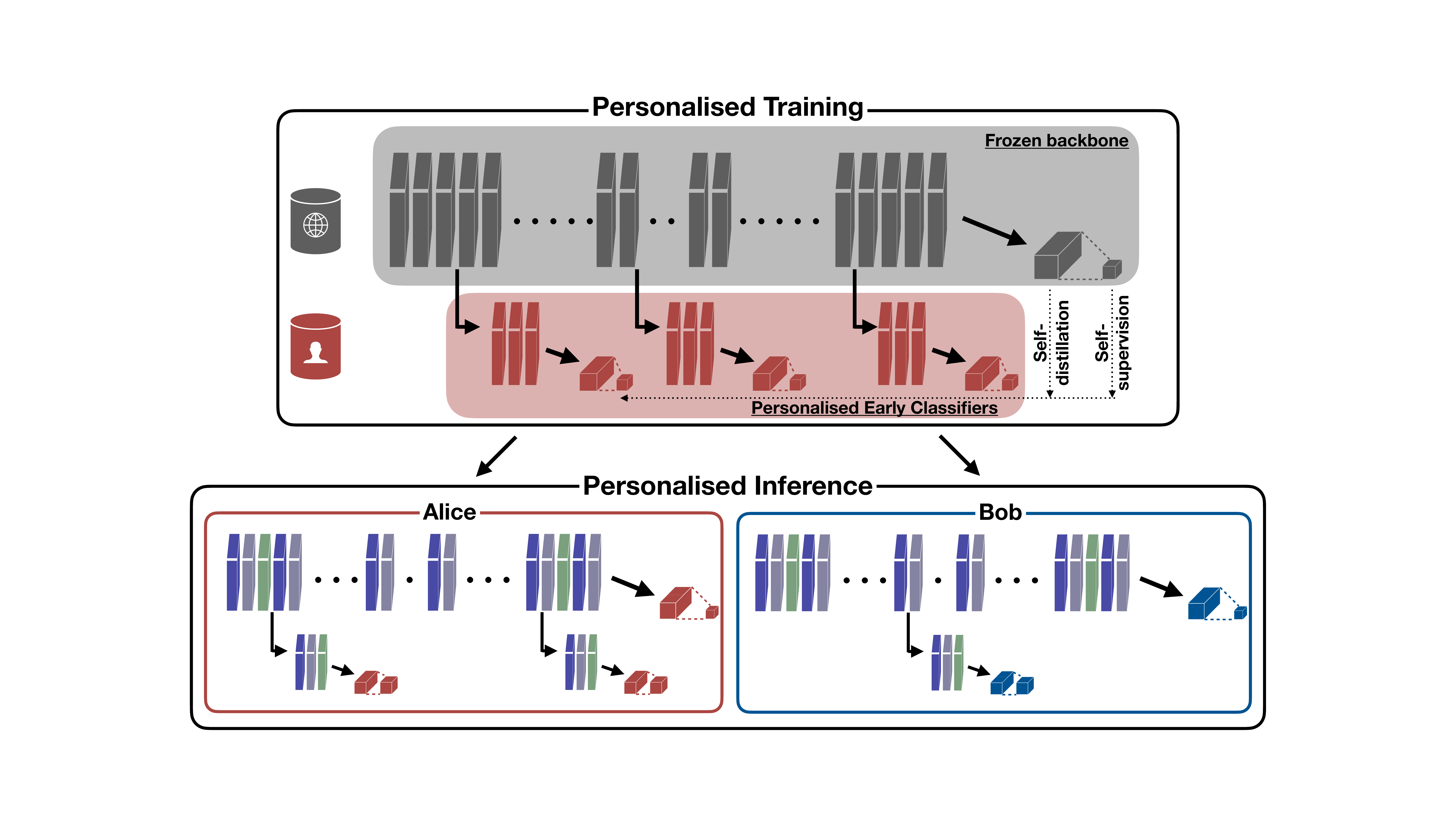}
      \vspace{-0.4cm}
      \caption{Overview of early-exit personalisation.}
      \vspace{-1em}
      \label{fig:ee-workflow}
\end{figure}

The recent progress of deep learning has enabled the development and ubiquitous deployment of numerous novel systems and apps.
At the same time, mobile chipsets are increasingly equipped with dedicated processing units (\textit{e.g.} NPUs) for efficient execution of DNNs~\cite{embench_2019,ai_benchmark_2019}. As such, on-device execution is becoming an emerging approach for meeting the latency, energy and privacy requirements of such systems.

% The recent radical progress in the field of deep learning has enabled the development of numerous novel systems and applications and their ubiquitous deployment. 
% %From autonomous robots to augmented reality apps~\cite{edge_ar2019mobicom}, deep neural networks (DNNs) have become a core component powering critical system-level tasks such as object detection, scene understanding and speech recognition. 
% At the same time, mobile-class chipsets are increasingly equipped with dedicated neural processing units (NPUs) for the efficient execution of DNNs~\cite{embench_2019,ai_benchmark_2019}. As a result, on-device execution is becoming an emerging approach of meeting the stringent latency, energy and privacy requirements of such systems.

Typically, deep models are trained centrally on standard datasets with the aim to generalise across samples and users. % upon deployment. 
For instance, facial landmark detectors are trained to capture various demographics, speech recognisers to accommodate %vastly 
different accents and voices, and home-assistant robots to work reliably across diverse household configurations. However, with data not being independent and identically distributed (IID) across devices in reality, the global model results in varying accuracy across the samples encountered in the wild, often failing catastrophically on unexpected inputs. 

\blue{Conventional solutions typically rely on cloud- or edge-based setups to perform \textit{model personalisation}. Under this scheme, each device transmits the user-specific data to a remote server, where personalisation takes place through additional training rounds. Although this approach takes advantage of the resource-rich server, it comes with significant overheads. First, the exposure of both the user data and the resulting personalised model to the remote side raises privacy concerns~\cite{cloud_privacy2011commsacm,edge_privacy2016jiot,occlumency2019mobicom,darknetz2020mobisys}. Second, on the service-provider side, using cloud/edge infrastructure comes with significant operating costs~\cite{talukder2010cloud}, due to the high demand of DNN training workloads for compute, memory and bandwidth resources.}

A promising approach to remedy this situation is \textit{on-device model personalisation}, aiming to tailor the DNN to each individual user or context. 
%Personalisation methods typically consist of two stages. 
%Initially, a global model is built and trained in a centralised manner. Next, the global model is personalised for each user using the client's private data. To enable fast adaptation, minimise server-side cost and comply with privacy regulations, 
% \steve{TODO: Citation here, can be GDPR related or sth}, 
%\textit{on-device personalised learning} can be a key solution.
Despite the algorithmic progress, there are still important limitations. First, the excessive resource demands of DNN training and the device hardware heterogeneity makes on-device personalisation challenging. Second, user data are frequently unlabelled, making supervised learning impossible. Third, personalisation can result in undetected catastrophic forgetting for non-frequent inputs. % outside of the personalised distributions. 

%To mitigate this, there is a need for novel methods that enable on-device personalisation without user data ever leaving the device, while simultaneously offering computational efficiency and adaptability to targets of varying capabilities.
% 
% while simultaneously offering computational efficiency and elasticity.
% Efficiency would secure that all processing steps of on-device training and inference impose low overhead and fully utilise the target device's capabilities. 
% Elasticity ensures the adaptability of the system to the availability of resources of the device upon deployment (i.e. capabilities) and at runtime (i.e. load).
% Elasticity would ensure the ability to sustain or improve performance in case of an increase in the amount of available resources.
% 
To address these limitations, we propose \tool, a novel framework that converts a pretrained CNN into a multi-exit network  and personalises its early exits to the specific user's data distribution (Fig.~\ref{fig:ee-workflow}).  Personalisation  takes place purely on-device and aims at producing classifiers along the depth of the network that are specialised for the user's data.
At inference time, the network can exit early if it is confident on its early output, or progressively refine the quality of the result.
Training can take place with or without ground-truth labels in a self-supervised manner, using the output of the network's last exit. By only personalising the early classifiers and keeping the backbone network frozen, we render the training process lightweight enough to take place overnight, while the device is plugged in. Finally, as the backbone is not altered, we can assess the quality of each personalised classifier and capture out-of-distribution samples at run time.

We evaluate our system across two  networks and datasets. Results indicate that we can achieve up to $3.2\times$ inference throughput gain, $3.1\times$ fewer FLOPs and $25.1\times$ fewer parameters while maintaining similar accuracy. Moreover, training an  exit can be up to $23\times$ faster than the whole network.

\vspace{-0.45cm}
\section{Background and Related Work}
\label{sec:related_work}
\vspace{-0.1cm}

Recently, an increasing body of work has focused on the design of early-exit networks, \textit{i.e.}~DNNs with intermediate classifiers along their depth that provide varying accuracy-latency trade-offs. 
The goal of this class of models is to offer adaptive accuracy-latency behaviour, either through an input-dependent execution with each sample stopping at the appropriate exit based on its difficulty or by extracting a subnetwork. 
Existing efforts span from hand-crafted early-exit models~\cite{msdnet2018iclr,scan2019neurips} to model-agnostic~\cite{branchynet2016icpr,sdn2019icml} and deployment-optimised frameworks~\cite{spinn2020mobicom,hapi2020iccad}. Focusing on transfer learning, REDA~\cite{reda2020mm} employs self-distillation to efficiently adapt earlier exits to tasks from different domains. 
While this line of work focuses on speeding up inference by training a single global model offline, \tool attempts to personalise exits in order to build a user-specific network that can save both computation and energy. \blue{Tangentially, \cite{hydranets2018cvpr} specialises CNNs by dynamically dropping layers based on \textit{offline} class clustering. Nevertheless, to repurpose it for personalisation, the number of users and the most common classes for each user need to be known \textit{a priori}, leading to poor scalability.}
% \stelios{HydraNets go here - dynamic routing (dropping blocks) based on class clustering. First, specialisation takes place offline. Second, to repurpose for personalisation, we need to know the number of users and the most common classes for each user. This leads to poor scalability.}

On the personalisation front, several methods have been proposed, leveraging user data to specialise generic pretrained models~\cite{delcroix2016context,Rajagopal_2020_CVPR_Workshops}. Currently, this space is dominated by meta-learning approaches~\cite{hospedales2020metalearning}. Such algorithms train a DNN so that it ends up with a good parameter initialisation that is amenable to rapid specialisation without requiring an excessive amount of data. Our approach, instead, uses personalisation as a way of progressively accelerating both on-device inference and training as more user-specific data become available. Orthogonal meta-learning techniques can be combined with \tool to further improve performance.
\blue{Resembling PersEPhonEE's self-supervised technique, another line of work employs online model distillation to obtain a user-adapted lightweight model~\cite{online_distill2019iccv}. The proposed two-model method is tailored for videos and requires the separate and frequent execution of both a costly teacher model and a lightweight student. Although this is suitable for high-end platforms, it is prohibitive for mobile devices.}
\section{System architecture}
\label{sec:arch}

\begin{figure}[t]
      \centering
      \vspace{-0.1cm}
      \includegraphics[trim={14cm 3cm 14cm 3cm},clip,width=0.70\columnwidth]{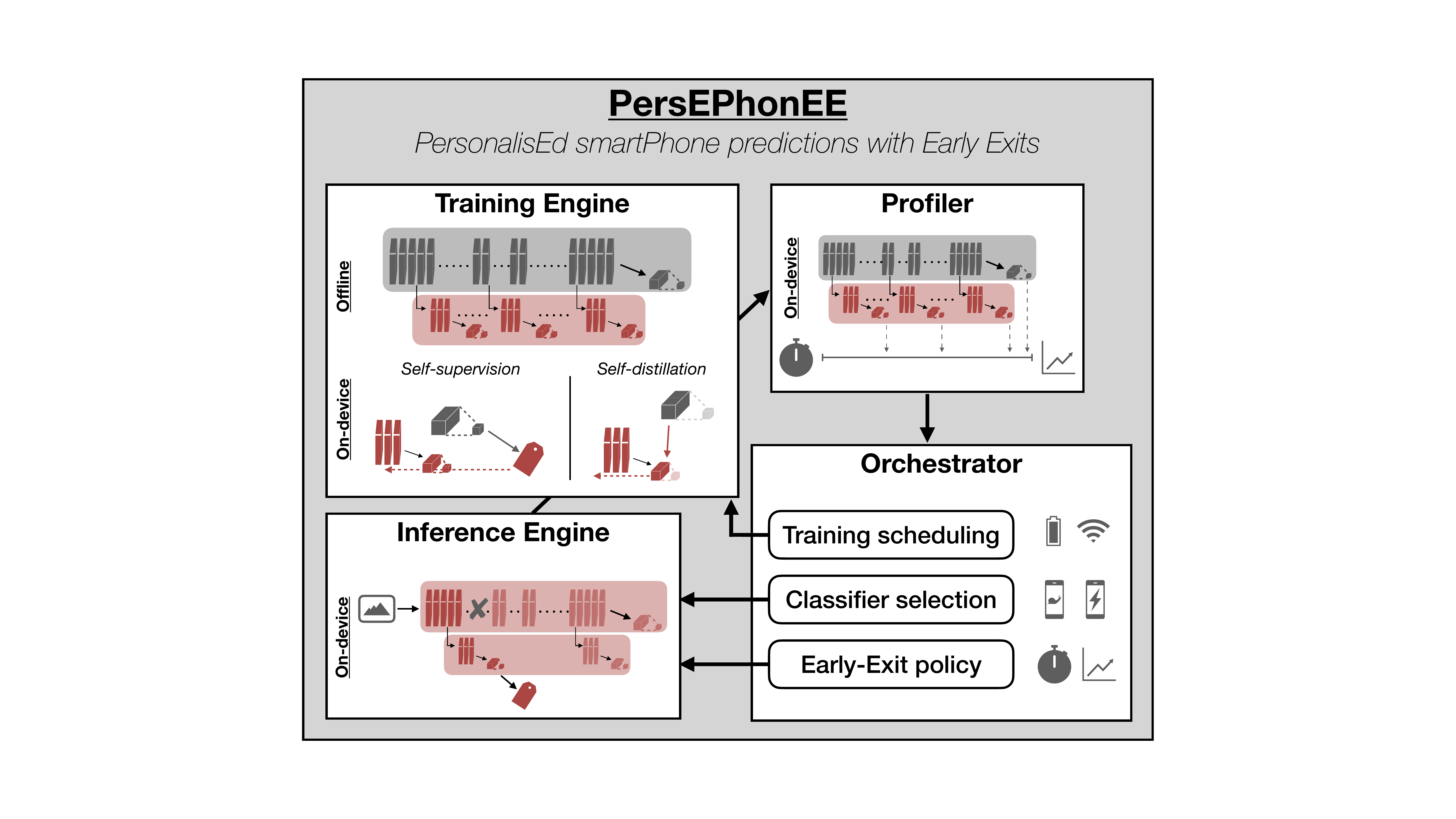}
      \vspace{-0.4cm}
      \caption{\tool's system architecture.}
    %   \vspace{-0.5cm}
      \vspace{-0.65cm}
      \label{fig:system-arch}
\end{figure}

% \il{There is some overlap with text from 3.1. Consider shrinking this section's intro }
\tool aims at creating an efficient personalised multi-exit model for each user. The processing flow (Fig.~\ref{fig:system-arch}) starts by attaching $M$ intermediate classifiers, or early exits, along the depth of a given CNN. 
% Depending on whether the backbone model has been pretrained or not, the \textit{Training Engine} applies either early-exit-only or joint backbone-exit training, respectively. 
Next, the \textit{Training Engine} trains both the early exits and the backbone network if it has not been pretrained.
Upon deployment to a user's device, the \textit{Profiler} collects statistics on the on-device execution. 
% and feeds them to the \textit{Orchestrator}. 
The \textit{Orchestrator} examines the \textit{Profiler}'s accumulated data and dictates when an on-device personalisation round will be launched.
%At this point, the \textit{Training Engine} performs on-device personalised training. 
% via either \textit{supervision}, \textit{self-supervision} or \textit{self-distillation}.
After a personalisation round, the \textit{Orchestrator} considers the accuracy-latency characteristics of each early exit and customises the execution by configuring the \textit{Inference Engine} with the selected exits and the early-exit policy to be used when processing new incoming data. 

\vspace{-1em}
\subsection{Training Engine}
\label{sec:trainer}

The \textit{Training Engine} comprises a dual structure: \textit{i)~}an \textit{offline} component (Sections~\ref{sec:multiexit_model} \& \ref{sec:global_train}) that derives a global multi-exit variant given a CNN and trains it using a generic dataset and \textit{ii)~}an \textit{on-device} component (Section~\ref{sec:personalisation}) that adapts the global multi-exit model to work well for the current user.

%Here we are to describe how we go from an untrained network to a multi-exit architecture with personalised classifiers. We can refer to a figure and then describe the pipeline of the system (i.e. how it works) briefly, and refer to subsections for the specifics.

\vspace{-0.2cm}
\subsubsection{Deriving a multi-exit model}
\label{sec:multiexit_model}

Given a CNN model, \tool converts it into a multi-exit network by attaching a number of early exits along its path. For the exit architecture, we utilise a uniform design for all early exits, adopting the intermediate classifier structure of MSDNet~\cite{msdnet2018iclr}.
Furthermore, we follow a platform-agnostic approach with equidistant placement of the $M$ classifiers along the depth of the backbone network in terms of FLOP count, \textit{i.e.}~at $i/(M+1)$-th, where $i\in[1,M]$ is the ordinal of the classifier.
% \textit{i.e.}~at $0.15 + 0/M$, $0.15+1/M$, ..., $0.15+(M-1)/M$ of the total FLOPs.

% There are a number of decisions to consider such as: {1)}~the \textit{number}, {2)}~the \textit{position} and {3)} the \textit{architecture}  of intermediate classifiers (early exits), 
% We place the intermediate classifiers along the depth of the architecture with equal distance in terms of FLOP count.  Early exits are allowed to be placed both after and within compound blocks, such as residual~\cite{He_2016} or Inception~\cite{Szegedy_2017} blocks. 
% With this platform-agnostic positioning strategy, we are able to obtain a progressive inference model that supports a wide range of latency budgets while being portable across devices. 
% With respect to their number, in our evaluation we use six early exits placed at 15\%, 30\%, \dots 90\% of the network's total FLOPs. Note that while more exits provide more opportunities for personalisation, they also result in larger model sizes. However, we can dynamically decide which exits to load and use on-device at runtime depending on the device capabilities and their perfomance after personalisation. 
% Last, we treat the architecture of the early exits as an invariant, adopting the design of~\cite{Huang2017}, so that all exits have the same expressivity~\cite{exp_icml_2017}. \il{Some text from SPINN here}

\vspace{-0.2cm}
\subsubsection{Training the global model}
\label{sec:global_train}

The multi-exit global model is then trained offline by utilising a generic training set for the target task.
If the supplied backbone network has been pretrained, we apply early-exit-only training, thus freezing the backbone's parameters and training only the intermediate classifiers' layers~\cite{hapi2020iccad}. If the supplied CNN is not trained, we jointly train the backbone and intermediate exits from scratch using the cost function introduced in~\cite{sdn2019icml}.
\blue{In terms of overhead, training from scratch the multi-exit model spans between 1.2$\times$-2.5$\times$ \steve{This is on the same server-grade h/w, right?}\stelios{Done.} the time of the backbone network, depending on the architecture and number of exits, with higher overhead when $\frac{FLOPs_{\text{early exit}}}{FLOPs_{\text{backbone}}}$ is larger. Nonetheless, as training takes place once for all users upfront, it is rapidly amortised by the runtime gains of confident samples from early exiting.}
% We jointly train all classifiers from scratch and employ the cost function introduced in~\cite{sdn2019icml} as follows: $\mathcal{L} = \sum_{i=0}^{N-1}{\tau_i*\mathcal{L}_i}$ 
% with $\tau_i$ starting uniformly at $0.01$ and linearly increasing it to a maximum value of $C_i$, which is the relative position of the classifier in the network ($C_0 = 0.15$, $C_1 = 0.3$, \dots, $C_{\text{final}}=1$). 
% 
% The rationale behind this is to address the problem of ``overthinking''~\cite{sdn2019icml}, where some samples can be correctly classified by early exits while being misclassified deeper on in the network. Typically, this results in a neural network where the accuracy is progressively increasing with respect to the position of the  exit, while the last exit maintains the accuracy of the original model. \stelios{The previous text in 3.3.2 is from SPINN.}
The resulting model is the trained \textit{global} model, serving as initialisation for 
% to be downloaded by all
client devices to personalise. The global model is then deployed to the mobile and embedded devices. 
% , similarly to how CNNs are currently deployed. 

% \steve{Question here: Have we tested how the model behaves on randomly initialised early-exits?}

\vspace{-0.2cm}
\subsubsection{On-device early-exit personalisation}
\label{sec:personalisation}

%While the goal of model personalisation is to adapt the generic global model to specific users, 
On-device personalisation is hindered by the limited  compute, storage and memory capabilities of mobile devices. Moreover, device diversity  calls for more elastic models that can adapt computation and energy usage.
%First, unlike server-based training, it is not feasible to store a large amount of data on embedded platforms.
At the same time, obtaining ground-truth labels for user data is often impractical. Finally,  prolonged fine-tuning can lead to catastrophic forgetting.
% \footnote{\emph{Catastrophic forgetting} occurs when a NN ``forgets'' prior knowledge backpropagated previously in training.}.

To overcome these  limitations, \tool introduces %multi-exit networks coupled with a tailored personalisation training algorithm. 
%The key idea behind the proposed approach is to introduce 
personalised multi-exit networks and an efficient 
% \il{I did some shrinking here as there was too much repetition and we need space}
%which require only a fraction of the resources, alleviating the computational and memory challenges of both training and inference. 
% if we were to personalise them only for a subset of the input distribution (\textit{e.g.}~a given user, device or environment).
% Our work leverages the fact that neural networks might require a fraction of the model parameters if we were to personalise them only for a subset of the input distribution (\textit{e.g.}~a given user, device or environment).
  on-device training scheme. We adopt a \textit{frozen-backbone} training approach that  updates the  parameters of the early exits only (Fig~\ref{fig:system-arch}). This strategy has a twofold gain: First, early exits and their gradients only occupy a fraction of the global model's memory usage and require a significantly lower number of FLOPs at training time. Second, as the network path to the last exit remains unmodified, the original output of the global model is maintained as a fail-safe, partly counteracting catastrophic forgetting by allowing for situations where the input sample does not follow the user-specific distribution.

% First, as fine-tuning the whole network is often prohibitive due to the excessive resource demands, 
% As a first step, we \textit{freeze} the backbone of the global model, including the original last exit and fine-tune only the pretrained early exits on-device. This strategy has a twofold gain. 
% First, fine-tuning the whole network with on-device resources is often infeasible due to the excessive resource demands. Instead, early exits only occupy a fraction of the global model's memory usage and require a significantly lower number of FLOPs.
% Secondly, we want to maintain the original output of the global model as a fail-safe, to allow for situations where the input does not follow the personalised distribution.

% 
% To achieve this, we \emph{freeze} the backbone of the global model, including the original -- last -- exit while we  fine-tune the pretrained early exits on-device.
% We only train the early exits for two reasons. First, fine-tuning the whole network on-device might be infeasible due to the computational and memory requirements;  early exits only have a fraction of the total parameters. Secondly, we want to maintain the original output of the global model as a fail-safe, to allow for situations where the input does not follow the personalised distribution.

To remedy the shortage of ground-truth labels in realistic scenarios, we design an objective function that personalises the multi-exit network using: \textit{i)~}supervision, \textit{ii)~}self-supervision or \textit{iii)~}self-distillation. 
% In a typical deployment, the user-specific input samples are used overnight to fine-tune the exits. 
% In most cases, we expect that the ground-truth labels might not be available. 
% For this reason, \tool personalises the early exits using: \textit{i)~}a supervised, \textit{ii)~}self-supervised or \textit{iii)~}self-distilled manner. 
When no hard labels are available, either the soft labels (\textit{i.e.}~the softmax distribution) (\textit{self-distillation}) or the top-1 prediction (\textit{self-supervision}) from the last exit -- which is typically the most accurate classifier in the global dataset -- can be used to ``teach" the early exits. 
The intuition is that the early exits, despite having smaller learning capacity, can progressively approach the accuracy of the last classifier for the \textit{personalised} input (\textit{i.e.}~a subset of the input distribution). If ground-truth labels are available, we can leverage them (\textit{supervision}) to further fine-tune the exits and, thus, achieve even higher accuracy.

To this end, we introduce a hybrid loss function that combines the three training schemes with a tunable weighting. Specifically, we define the personalisation loss of exit $i$ as:
\begin{small}
\begin{flalign}
    \mathcal{L}^{(i)}_{\text{personal}} &  (\hat{\boldsymbol{y}}^{(i)}_T,\hat{\boldsymbol{y}}^{(M+1)}_T,\boldsymbol{y}) = & \\
    & excl(\alpha, \gamma) \cdot \mathcal{L}_{\text{superv}}^{(i)}\left(\hat{\boldsymbol{y}}^{(i)}_{T=1}, \boldsymbol{y}\right) \nonumber & \\ 
     &+ \beta \cdot \mathcal{L}_{\text{self-distill}}^{(i)}\left(\hat{\boldsymbol{y}}^{(i)}_T, \hat{\boldsymbol{y}}^{(M+1)}_T\right) \nonumber & \\
     &+ excl(\gamma, \alpha) \cdot \mathcal{L}_{\text{self-superv}}^{(i)}\left(\hat{\boldsymbol{y}}^{(i)}_{T=1}, \text{top1}(\hat{\boldsymbol{y}}^{(M+1)}_{T=1})\right) \nonumber & 
     \label{eq:exit_lostt}
\end{flalign}
\end{small}
% 
% \begin{equation}
%     \mathcal{L}^{(i)}_{\text{personal}}(\hat{\boldsymbol{y}}^{(i)}_T,\hat{\boldsymbol{y}}^{(M+1)}_T,\boldsymbol{y}) = \alpha \cdot \mathcal{L}_{\text{soft}_{T}}^{(i)} + (1 - \alpha) \cdot \mathcal{L}_{\text{hard}}^{(i)}
%     \label{eq:exit_lostt}
% \end{equation}
% 
where {\small $\mathcal{L}_{\text{superv}}^{(i)}$} is the cross-entropy loss between the hard label, $\boldsymbol{y}$, and the output of the exit $i$, $\hat{\boldsymbol{y}}^{(i)}_{T=1}$, {\small $\mathcal{L}_{\text{self-distill}}^{(i)}$} is the KL divergence between the output of exit $i$, $\hat{\boldsymbol{y}}^{(i)}_T$, and the last exit, $\hat{\boldsymbol{y}}^{(M+1)}_T$, smoothed by temperature $T$~\cite{hinton2014distilling}, {\small  $\mathcal{L}_{\text{self-superv}}^{(i)}$} is the cross-entropy loss treating the last exit's top-1 prediction as the ground-truth label, and $excl(x,y) = (x=0 \lor y=0)$ ensures that $x$ and $y$ are mutually exclusive. 
Hyperparameters $\alpha$, $\beta$ and $\gamma$ determine the importance of the three components, weighing the supervised, self-distillation and self-supervised loss, respectively. By replacing cross-entropy with another loss, the \textit{Trainer} can also target other tasks such as regression. 
Finally, the overall training objective is to minimise the sum of losses of all $M$ exits across the user-specific data.
% For the samples without hard labels we use $\alpha = 1$, \textit{i.e.}~only soft labels are used. \stelios{This is a special case. We can just specify the values used in the Evaluation.} Notice that fine-tuning the early exits typically requires a small number of epochs with small learning rate as the exits were already pretrained during the multi-exit global model training stage. \steve{We want to refine, not find a new global optimum. Is that a valid hypothesis?}

% The overall training objective is to minimise the loss across all exits and can be formulated as
% \begin{equation}
%     \min \sum\limits_{i=1}^M \mathbb{E}_{(\boldsymbol{x},\boldsymbol{y})\in \mathcal{D}} \mathcal{L}_{\text{personal}}^{(i)}(\hat{\boldsymbol{y}}^{(i)}_T,\hat{\boldsymbol{y}}^{(M+1)}_T,\boldsymbol{y})
%     \label{eq:total_loss}
% \end{equation}
% \stelios{The equation above is a bit strange. Not sure how to express the expectation because not all data have labels.}

\vspace{-0.2cm}
\subsubsection{Confidence calibration}
\label{sec:conf_calibr}

For the early-exit policy, we estimate a classifier's \textit{confidence} for a given input using the top-1 output value of its softmax layer~\cite{calibration2017icml}, defined as {\small $\text{softmax}(z)_i = \frac{e^{z_i}}{\sum_{j=1}^{K}e^{z_j}}$} for the i-th classifier. 
% (Eq.~(\ref{eq:conf_threshold})).
% (Eq.~(\ref{eq:softmax}))~\cite{calibration2017icml}.
An input takes the i-th exit if the prediction confidence is higher than a tunable threshold, $thr_{\text{conf}}$, following {\small $\arg_i\{\max_i\{\text{softmax}_i\} > {thr}_{\text{conf}}\}$}.
The exact value of $thr_{\text{conf}}$ provides a trade-off between the latency and accuracy of the multi-exit model and determines the \textit{early-exit policy}. If none of the classifiers reaches the confidence threshold the original (last) classifier is used as a fail-safe for non-personalised inputs. 

% \vspace{-0.8em}
% \begin{small}
%     \begin{align}
%         &\text{softmax}(z)_i = \frac{e^{z_i}}{\sum_{j=1}^{K}e^{z_j}} & ~~~\text{ (\textit{i-th classifier softmax})} \label{eq:softmax}\\
%         &\arg_i\{\max_i\{\text{softmax}_i\} > {thr}_{\text{conf}}\} & ~~~\text{ (\textit{Confidence check})} \label{eq:conf_threshold} %\\
%         % &\argmax_{j\in\text{classifiers}}\{\max_i\{\text{softmax}_i^j\}\} & ~~~\text{ (\textit{Return most confident})}  \label{eq:max_conf}
%     \end{align}
% \end{small}
% \vspace{-1.1em}

\tool re-calibrates its confidence threshold after every personalisation round. This is performed by maintaining a calibration set, obtained from the user's input distribution. After the early-exit personalisation, the calibration set is used to measure the average loss of each exit with respect to the original model output, and the exit rate of each exit which will affect the overall average inference latency. With these measurements, \tool performs a Pareto-front analysis in the accuracy-average latency space and chooses the smallest threshold that maintains the model's accuracy within a specific drop tolerance. 
Furthermore, early exits that do not contribute much to the user-specific distribution (\textit{e.g.}~they have a high loss) are pruned. 
% \steve{Validation --> calibration and define size to be efficiently deployed. Probably this calibration should have some examples explicitly not in the training set to estimate the out-of-distribution confidence of the early classifier (we need this to be low)}

\vspace{-.8em}
\subsection{Inference Engine}
\label{sec:inf_engine}
% \il{I would move the inference engine last ...} \stelios{But it's controlled by Orchestrator. It'd be good to have it introduced before.}\steve{Agree with Stelios I think on that. If anything, I would put it close to the trainer.} \stelios{Agreed, let's move it before the Profiler.}
The \textit{Inference Engine} is responsible for executing the forward pass of the multi-exit network over a supplied set of input samples. These are either incoming samples to be classified or stored past samples to be used for personalisation. The \textit{Orchestrator} configures the \textit{Inference Engine} by defining which early exits to use and the early-exit policy, \textit{i.e.}~the threshold above which an exit's prediction is considered confident.

\vspace{-1em}
\subsection{Profiler}
\label{sec:profiler}

In order for \tool to make informed decisions about the cost (\textit{i.e.}~latency, memory usage) and benefits (\textit{i.e.}~accuracy) of using each classifier, we need to measure these metrics for a personalised model on the target device. For this reason, the \textit{Profiler} gets invoked after a training session has completed, while the phone is still plugged in. Specifically, the profiler conducts one forward pass of the calibration set, estimating \textit{1)~}the latency, \textit{2)~}top confidence and \textit{3)~}accuracy of each classifier ($i$), using the final prediction as the ground-truth if no hard labels are present. 
Subsequently, these metrics are passed to the \textit{Orchestrator} (Section~\ref{sec:orchestrator}) to decide the early-exit policy during inference.

% \begin{small}
% \vspace{-1.1em}
% \begin{equation}
%     \text{profiler}: \begin{cases}
%         lat_{i} &\text{(\textit{latency})} \\
%         mem_{i} &\text{(\textit{memory usage})}\\
%         \text{max}\{\text{softmax}(logits_i)\} &\text{(\textit{confidence})} \\
%         acc_{i} &\text{(\textit{accuracy})} \\
%     \end{cases}
% \end{equation}
% \end{small}
% 

\vspace{-1em}
\subsection{Orchestrator}
\label{sec:orchestrator}
\blue{
The \textit{Orchestrator} is a key system component that measures the run-time performance and is responsible for scheduling the different operating phases of \tool and configuring the \textit{Inference Engine}.}
Three basic phrases are defined: 

% \textit{i)}~inference, \textit{ii)}~background exploration and \textit{iii)}~overnight personalisation.
\vspace{-0.1cm}
\begin{itemize}[leftmargin=0em,label={}]
    \item \textbf{Inference phase}: The selected exits and threshold are used to configure the \textit{Inference Engine}, saving computation time and energy on the new samples.
    % Furthermore, the sample is stored on-device to later be used in the personalisation phase. 
    % If the input is large (\textit{e.g.}~video frames) a random sample can be stored. \steve{This is sloppy. Why would a random sample of a video be useful? This is very much task dependent.}
    \item \textbf{Exploration phase}: \tool periodically re-evaluates the quality of the early exits in the background, with a given probability $p_{\text{expl}}$. This is achieved by stochastically extracting the output of the last classifier and calculating the loss function for each of the selected early exits. If the average loss of an intermediate classifier is consistently higher, there has probably been a domain shift. To mitigate this, we raise the confidence threshold to $thr^\prime_{\text{conf}}$ and schedule the trainer to run on the newly encountered samples when the device is plugged in. The values of $p_{\text{expl}}$ and $thr^\prime_{\text{conf}}$ can be selected based on the transient load and battery level of the device as well as the loss increase of the early classifier.
    % In this manner, a potential shift in the user distribution is detected and the threshold is re-adjusted accordingly to $thr^\prime_{\text{conf}}$, in a similar manner to confidence calibration (Section~\ref{sec:conf_calibr}). \steve{This is very high level and we might have trouble from the reviewers.} \stelios{I'll rewrite.}
    \item \textbf{Personalisation phase}: The on-device personalisation task is scheduled overnight~\cite{sysdesign_fl2019mlsys} and %, when the device is typically charging, 
    only when there are sufficient new data on the device (Section~\ref{sec:samples_required}) or when the active threshold (Section~\ref{sec:ondevice_perf_eval}) $thr^\prime_{\text{conf}}$ deviates significantly from the validation set threshold $thr_{\text{conf}}$, as this indicates that the user input distribution might have shifted since the last time the exits were trained. 
    \blue{Given the aforementioned conditions, personalisation typically takes place on infrequent intervals, \textit{e.g.}~monthly, and only after the device is plugged to a power adapter and fully charged. Hence, the personalisation energy cost is minimal compared to the resulting inference energy savings during normal usage, \textit{i.e.}~when on battery power.}
\end{itemize}

% \vspace{-1em}
\section{Evaluation}
\label{sec:eval}

We have developed \tool on top of \textit{PyTorch} 
% (1.1.0) 
and \textit{torchvision} 
% (0.3.0)
and targeted two CNNs, namely ResNet-50~\cite{He_2016} and MobileNetV2~\cite{Sandler}.
% , with each model representing a different CNN architecture design. 
\blue{All reported measurements are taken on an Nvidia Jetson Xavier board equipped with a Quad-core Arm Cortex-A57 and a dual-core NVIDIA Denver2 CPU. We only use the embedded CPUs as they are comparable to today's flagship phones.}
% \steve{I wouldn't say that. One is 10W tops, the other is 30W.} \stelios{That's true :facepalm. Although when using only the CPUs, I doubt it would reach 30W.}
% 
For our experiments, we attach six early exits ($M=6$) in addition to the original final classifier, giving seven total possible output locations. We pretrain the global model (including the exits) on the global labelled dataset in an offline manner and we only personalise early classifiers on-device with or without the presence of ground-truth labels, leaving the final output untouched. We fine-tune the exits for 10 epochs with learning rate $0.01$.
% \stelios{We should mention very briefly sth about the training hyperparams.} \steve{This is gonna take space to explain for EMNIST and MobileNet I guess.}

\vspace{-1em}
\subsection{Datasets}
To demonstrate the ability of early exits to personalise we evaluate \tool on two different datasets:
% \begin{itemize}[leftmargin=1.1em,]
\begin{itemize}[topsep=0pt,leftmargin=0em,label={}]
    \item {\bf Personalised ImageNet:} Based on ILSVRC 2012 \cite{deng2009imagenet}, this is a personalised variant created with the assumption that not all users experience all 1000 labels with the same frequency. To this end, each personalised user is assumed to be exposed to images with labels whose frequency follow a Gaussian distribution to better capture such biases \cite{bias1}. After generating the user-specific label popularity, we sample images from ImageNet to create each user's personalised dataset.
    % training, validation and test sets. 
    \item {\bf FEMNIST:} This is a real-world dataset aimed at hand-written text recognition~\cite{cohen2017emnist}. It contains 80K 28$\times$28 images from 3.5K users (225 images per user) belonging to 62 different classes (10 digits, 26 lowercase, 26 uppercase letters). 
    We mix the data from 3250 random users to train the global model and evaluate the personalisation results on each of the remaining 250 users, holding out 100 images per user for testing.
\end{itemize}

\vspace{-1em}
\subsection{Accuracy on Personalised Exits}

\begin{figure}[t]
\centering
\vspace{-0.2cm}
    \begin{subfigure}[t]{0.49\columnwidth}
          \centering
          %\vspace{-0.5cm}
          \includegraphics[width=\columnwidth]{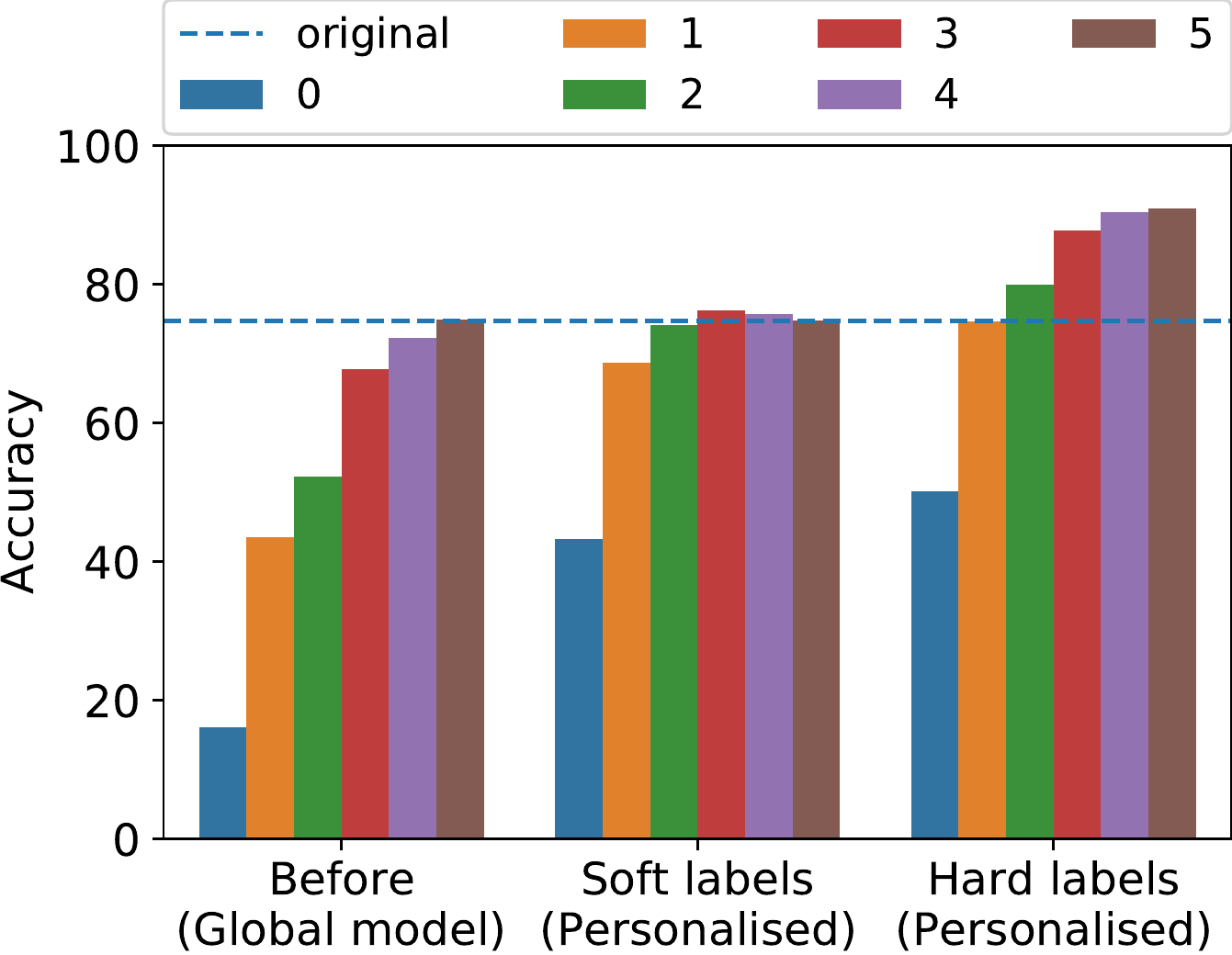}
          \caption{Personalised ResNet-50 on ImageNet. }
        %   \vspace{-0.5cm}
          \label{fig:imageNet}
    \end{subfigure}
\hfill
    \begin{subfigure}[t]{0.49\columnwidth}
          \centering
          %\vspace{-0.5cm}
          \includegraphics[width=\columnwidth]{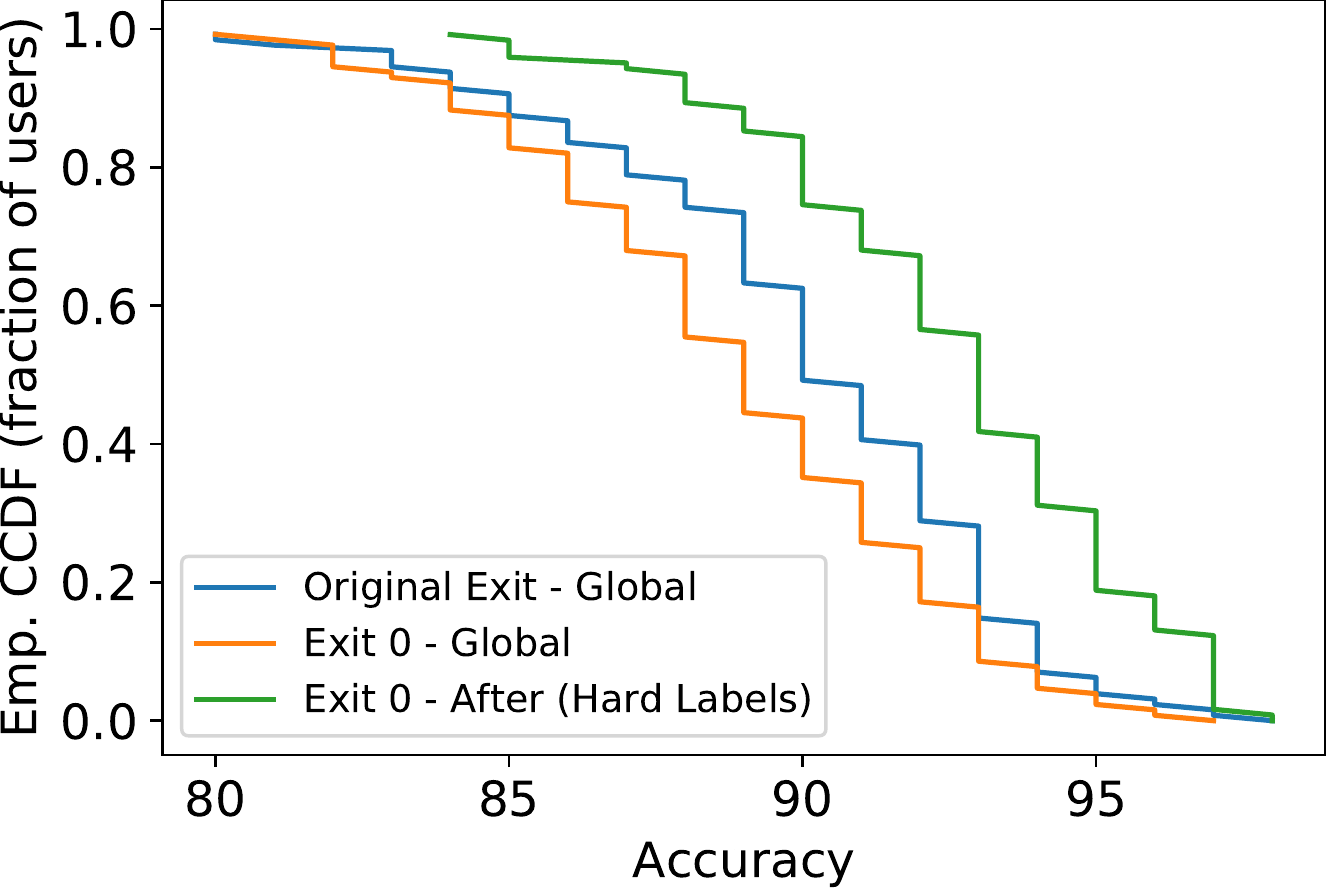}
          \caption{Exit-0 of personalised \mbox{MobileNetV2} on FEMNIST.}
        %   \vspace{-0.5cm}
          \label{fig:mobilenetv2}
    \end{subfigure}
    \vspace{-0.3cm}
    \caption{Accuracies per early classifier.}
    \label{fig:accuracies_ee}
    \vspace{-0.6cm}
\end{figure}

First, we evaluate the impact of on-device personalisation on the accuracy of individual early exits, using ResNet-50 on ImageNet and MobileNetV2 on FEMNIST (Fig.~\ref{fig:accuracies_ee}). 

Fig.~\ref{fig:imageNet}-Before shows the accuracy of the \textit{pretrained} exits on the personalised ImageNet dataset, whereas the dotted line shows the original model's accuracy. As expected, the earlier the exits are attached to the backbone, the lower the accuracy is, ranging from 16\% for exit-0 to 74.95\% for exit-5, similar to the original top-1 accuracy of 74.96\%. 
By personalising the early exits on-device with user-specific input, we can greatly improve their accuracy for similar (personalised) inputs. 
Under the presence of ground-truth labels (Fig.~\ref{fig:imageNet}-Hard Labels), \tool applies supervised personalisation with exit-1 reaching the original accuracy and pruning a large part of the computation. Later exits attain even higher accuracy than the global model (up to 90.9\% for exit-5) as they are now specialised for the user's frequently seen labels. 
When no labels are available on-device, \tool uses self-distillation. 
% (\textit{i.e.}~self-distillation where soft labels from the original last exit are used to fine-tune the early exits), 
In this case, although the early exits do not surpass the accuracy of the global model, some of the them still reach its accuracy significantly faster (\textit{e.g.}~exit-2).

Fig.~\ref{fig:mobilenetv2} depicts a different view, where we fix the personalised classifier and look into the empirical CDF of users reaching a specific accuracy. It can be observed that by using supervised personalisation, the same users can achieve higher accuracy faster. For instance, 75\% of the users attain 90\% accuracy at exit-0, while without personalisation this was for 40\% and 60\% for the non-personalised final exit.

\vspace{-1em}
\subsection{Training Samples Required}
\label{sec:samples_required}
\begin{figure}[t]
   \centering
   \vspace{-0.2cm}
    \begin{subfigure}{0.49\columnwidth}
      \centering
      \includegraphics[width=\columnwidth]{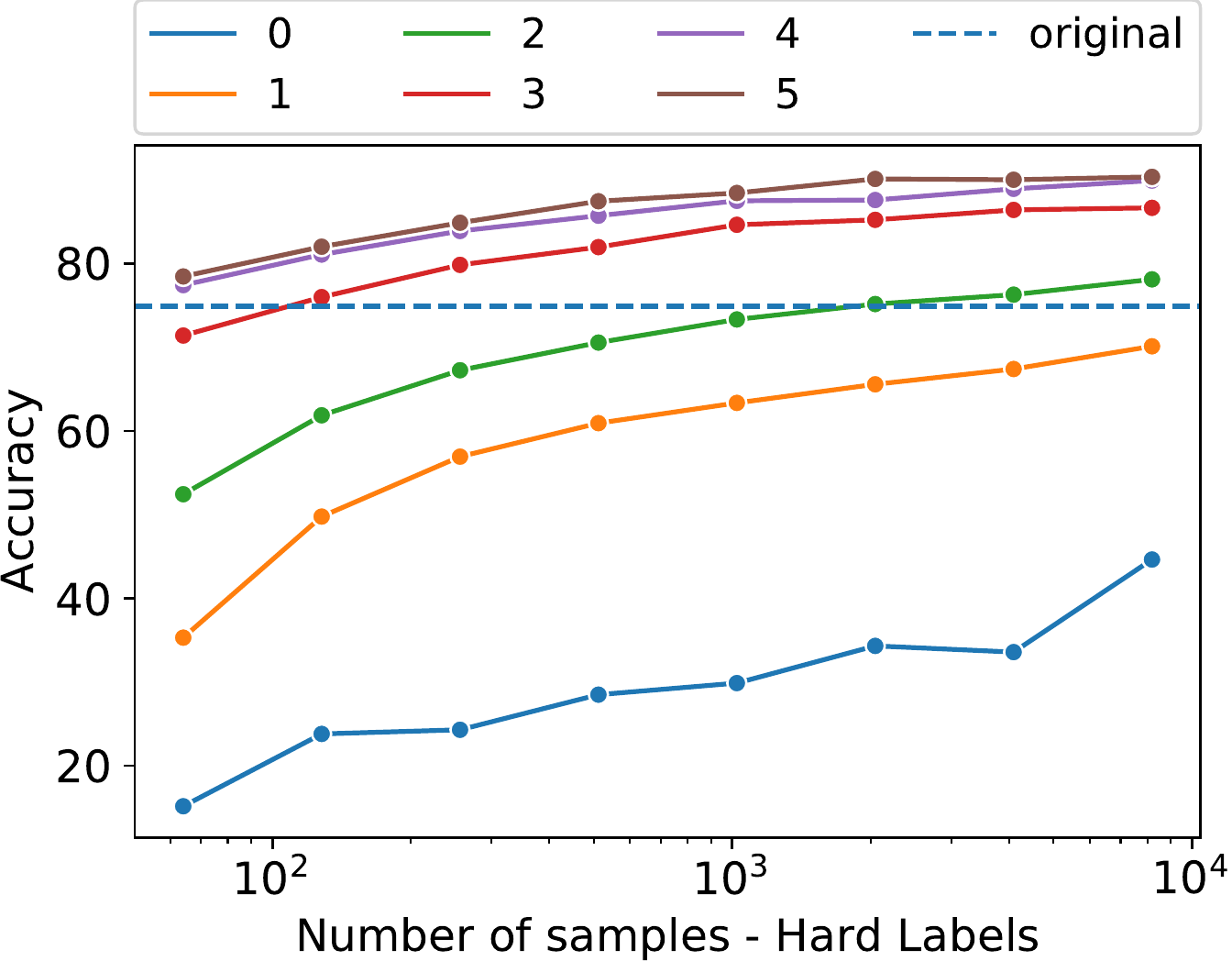}
      \caption{Hard labels}
      \label{fig:s-a}
    \end{subfigure}%
    \hfill
    \begin{subfigure}{0.49\columnwidth}
      \centering
      \includegraphics[width=\columnwidth]{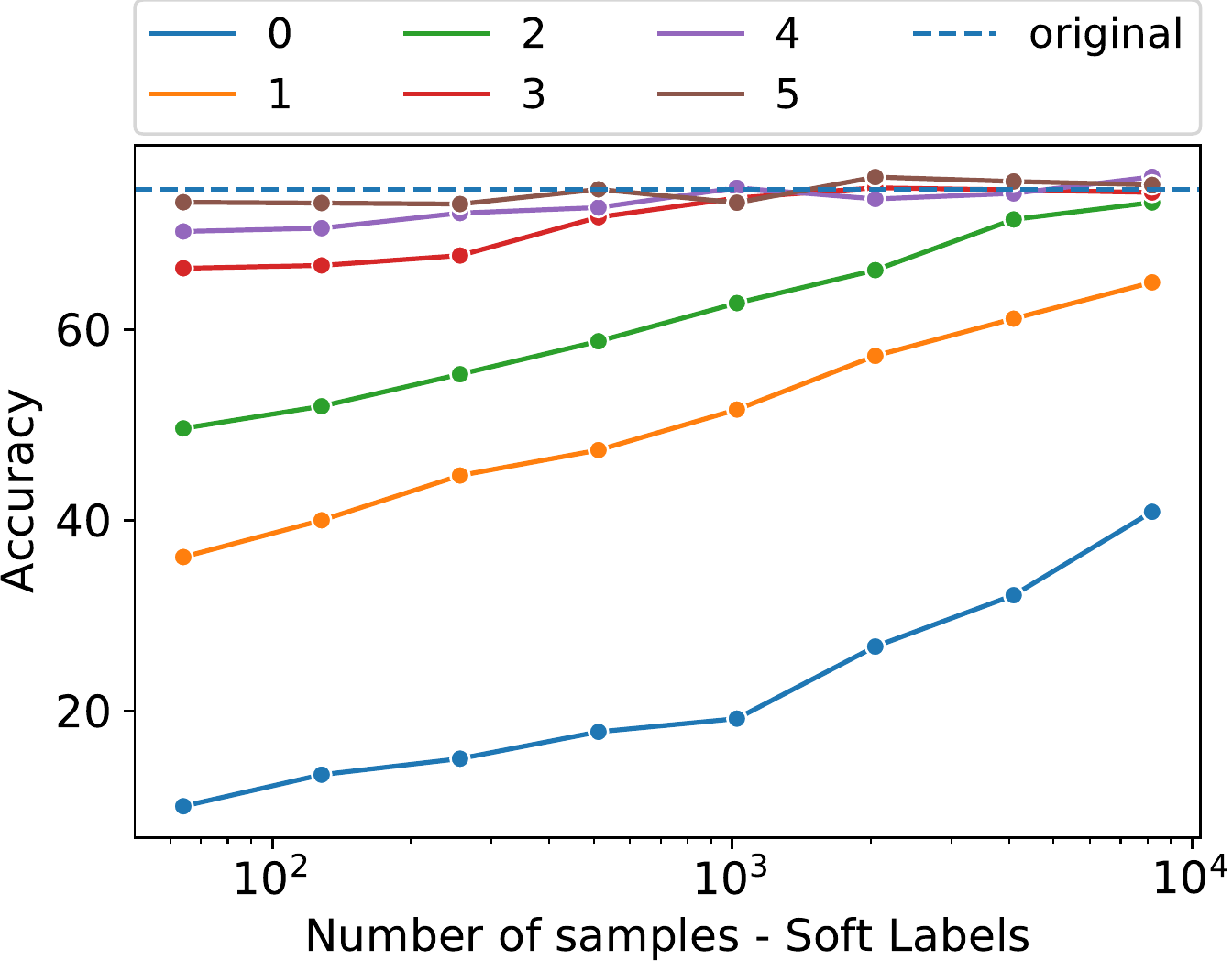}
      \caption{Self-distillation}
      \label{fig:s-b}
    \end{subfigure}%
    \vspace{-0.3cm}
    \caption{Accuracy vs. \#personalisation samples.}
    %   \vspace{-0.5cm}
    \label{fig:data}
    \vspace{-1em}
\end{figure}

% As mentioned in Section ~\ref{sec:arch},
Different users might need different early-exit policies as they might have varying amounts of data to train the model as well as different input complexity. Therefore, it is important to understand how much data are required to train a personalised multi-exit model. Fig.~\ref{fig:data}, depicts the accuracy achieved per early exit for varying number of samples.
% (from 64 to 8192). 
First, we note that the accuracy of every exit is gradually improving as more data become available. Moreover, the earlier the exit, the larger the improvement. As a result, different exits can be used for users with different amount of data. For example, users with 2K samples or more could utilise exit-2 to achieve similar performance to the original model, whereas users with solely 150 samples can use exit-3, still saving more than 60\% of the computational cost.

\vspace{-1em}
\subsection{Performance Evaluation}

\subsubsection {Performance - Accuracy Trade-off}

\begin{figure}[t]
    \centering
    \vspace{-0.15cm}
    \begin{subfigure}{0.49\columnwidth}
      \centering
      \includegraphics[width=\columnwidth]{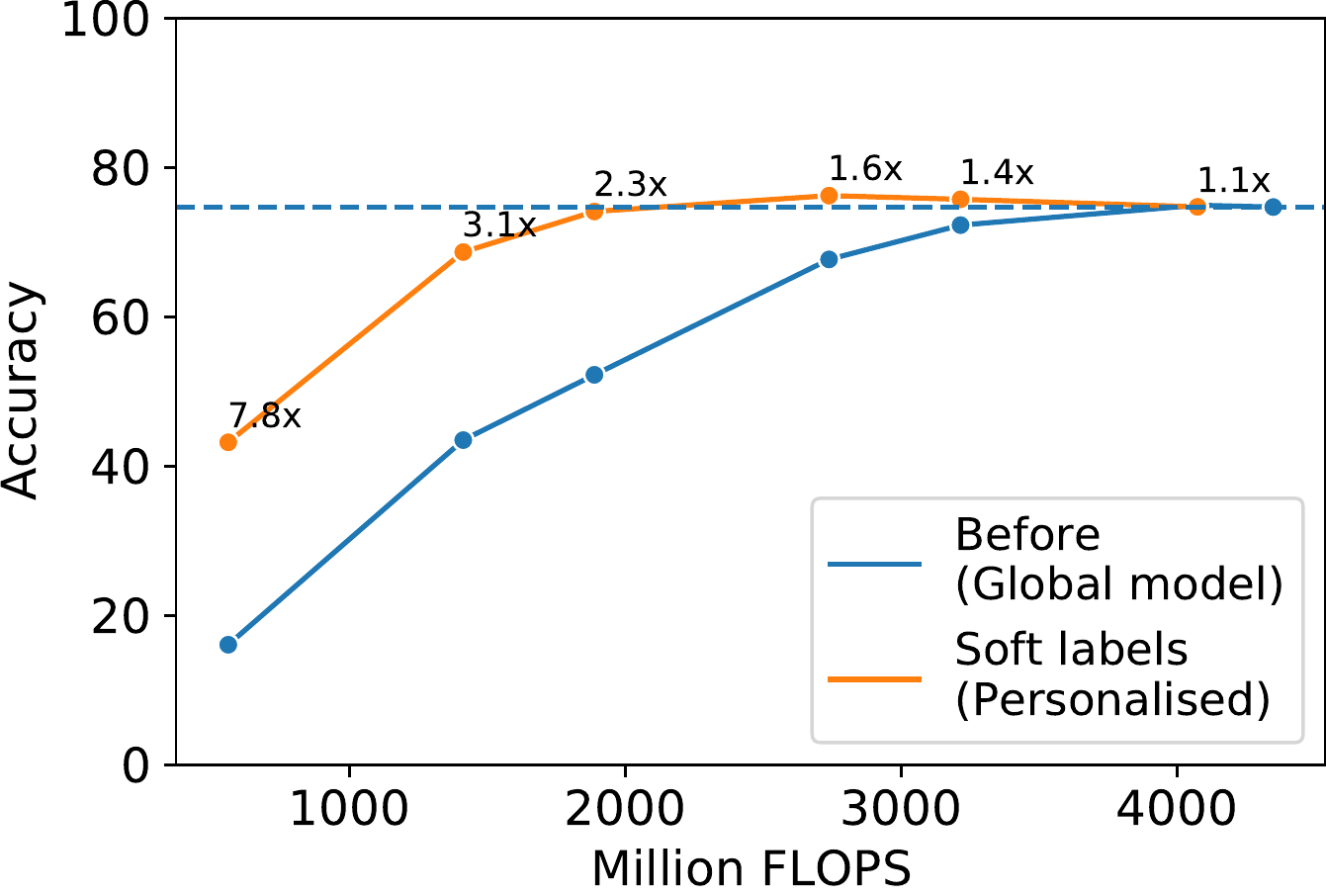}
      \caption{FLOPs (Self-distillation)}
      \label{fig:f-a}
    \end{subfigure}%
    \hfill
    \begin{subfigure}{0.49\columnwidth}
      \centering
      \includegraphics[width=\columnwidth]{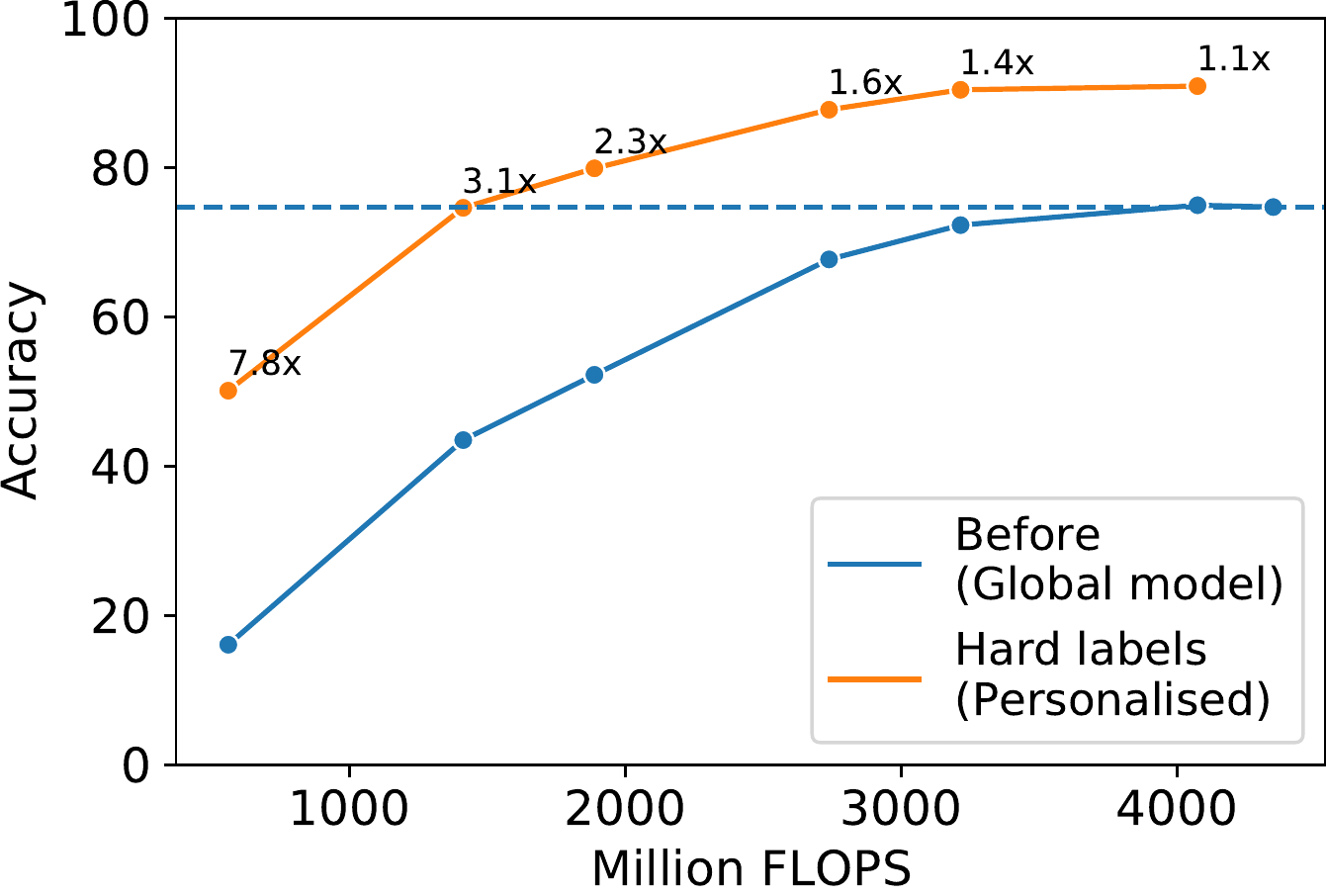}
      \caption{FLOPs (Hard labels)}
      \label{fig:f-b}
    \end{subfigure}%
    \hfill
    \begin{subfigure}{0.49\columnwidth}
      \centering
      \includegraphics[width=\columnwidth]{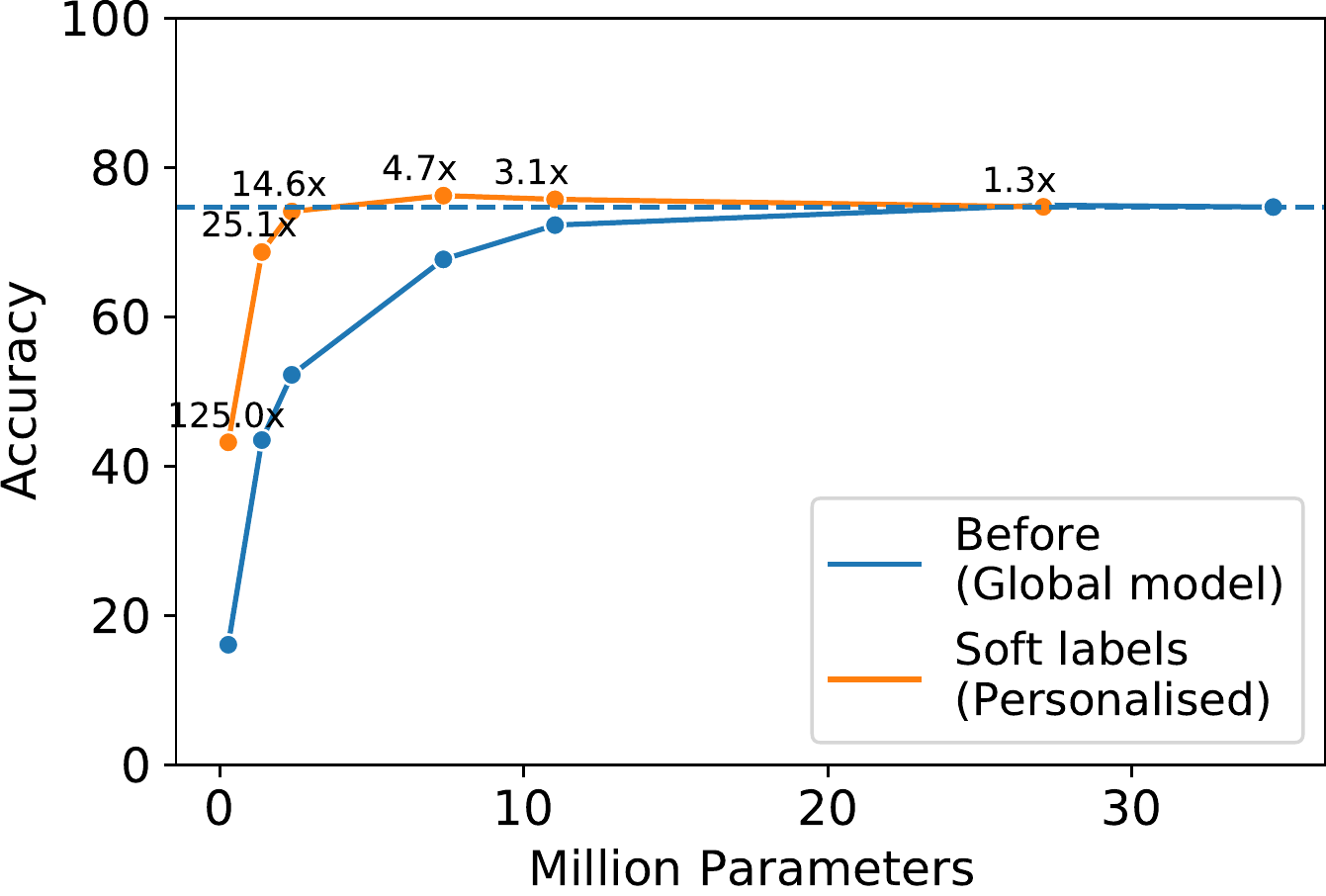}
      \caption{\mbox{Params. (Self-distillation)}}
      \label{fig:f-c}
    \end{subfigure}
    \hfill
     \begin{subfigure}{0.49\columnwidth}
      \centering
      \includegraphics[width=\columnwidth]{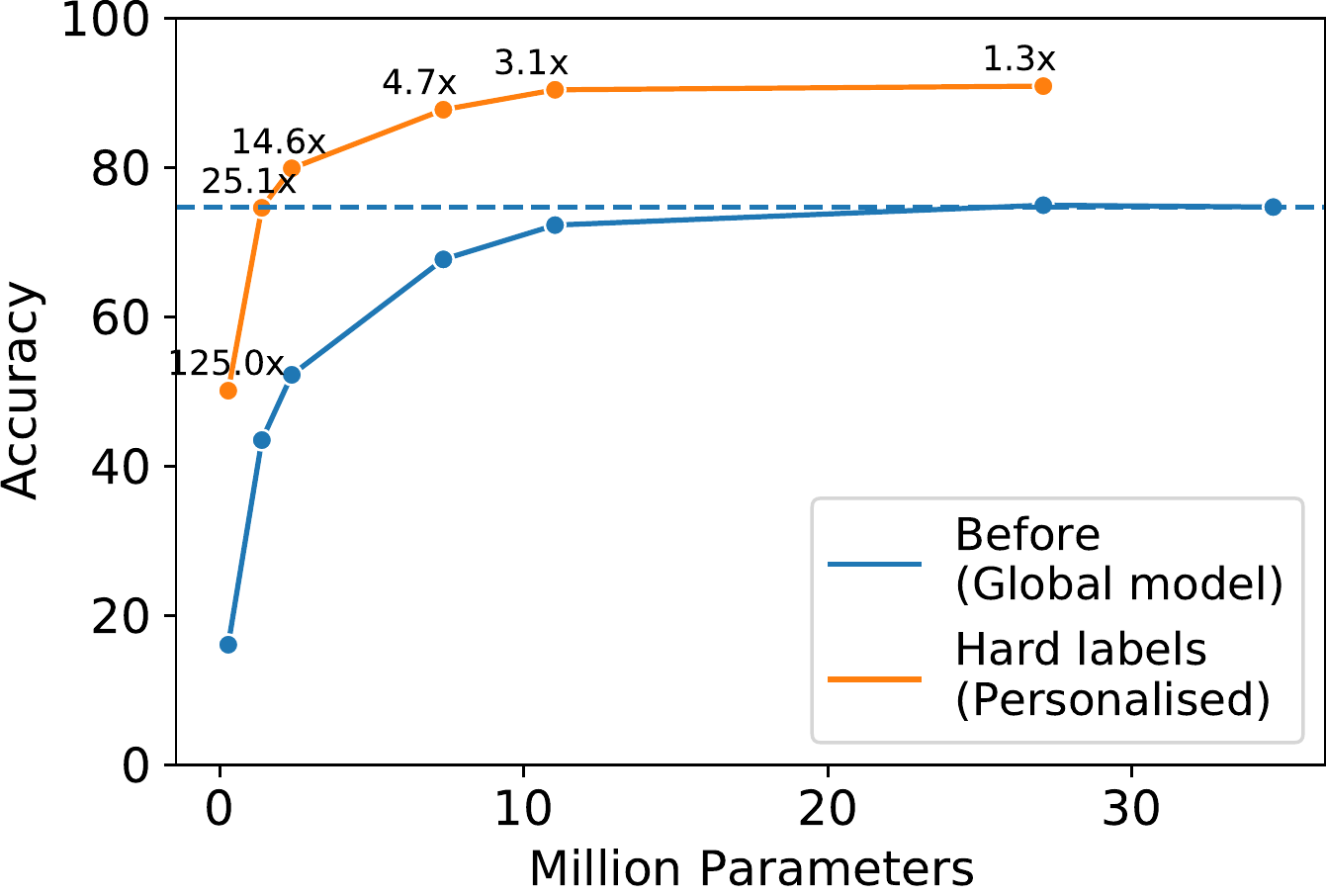}
      \caption{Params. (Hard labels)}
      \label{fig:f-d}
    \end{subfigure}%
    \hfill
    \vspace{-0.3cm}
    \caption{Accuracy vs. \#parameters and FLOPs for each exit, before and after personalisation.}
    \label{fig:flops-params}
    \vspace{-1.5em}
\end{figure}

% \il{Note that this figure contains all the info of figure 3 but it might be better to keep figure one to make it a bit easier to understand.}

Here, we correlate the achieved accuracy of our personalised early-exit model with the amount of computation (FLOPs) and memory (\#parameters) required.
In Fig.~\ref{fig:flops-params}, we depict this relationship for each of the exits.
With ground-truth (hard) labels, we observe that the personalised classifiers can reach the original accuracy with $25.1\times$ fewer parameters (Fig.~\ref{fig:f-b}) and $3.1\times$ fewer FLOPs (Fig.~\ref{fig:f-d}). Furthermore, accuracy is improved by 13 percentage points (pp) (to 87.8\%) while using $4.7\times$ fewer parameters and $1.6\times$ fewer FLOPs, resulting in a configuration that provides significant speedup, energy and accuracy gains. 
With no labels available, on-device self-distillation results in similar accuracy to the original network with $14.6\times$ fewer parameters (Fig.~\ref{fig:f-a}) and $2.3\times$ less FLOPs (Fig.~\ref{fig:f-d}), exploiting the more narrow, personalised input distribution. 

\vspace{-0.2cm}
\subsubsection{On-device performance}
\label{sec:ondevice_perf_eval}

% \steve{Should this section be near the FLOPs/params one? Maybe we can combine them in ``Performance Evaluation'' and have two subsections, one for training and one for inference.} 
In this section, we evaluate the training and inference performance on mobile CPUs to assess the feasibility of personalising CNNs on device.

\begin{figure}[t]
    \centering
    \vspace{-0.2cm}
    \begin{subfigure}{0.49\columnwidth}
      \centering
      \includegraphics[width=\columnwidth]{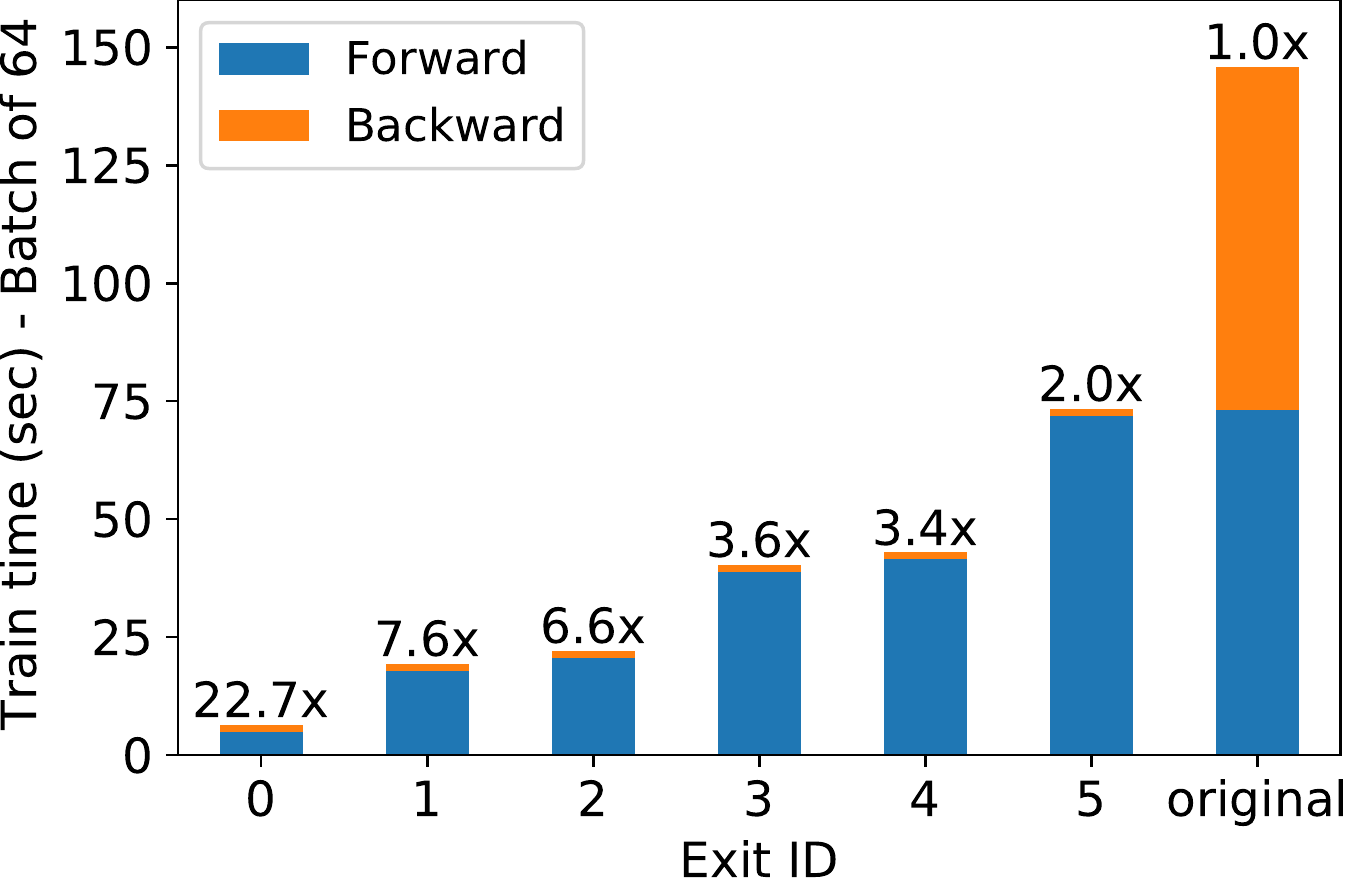}
      \caption{Per-exit training time.}
      \label{fig:t-b}
    \end{subfigure}%
    \hfill
    \begin{subfigure}{0.49\columnwidth}
      \centering
      \includegraphics[width=\columnwidth]{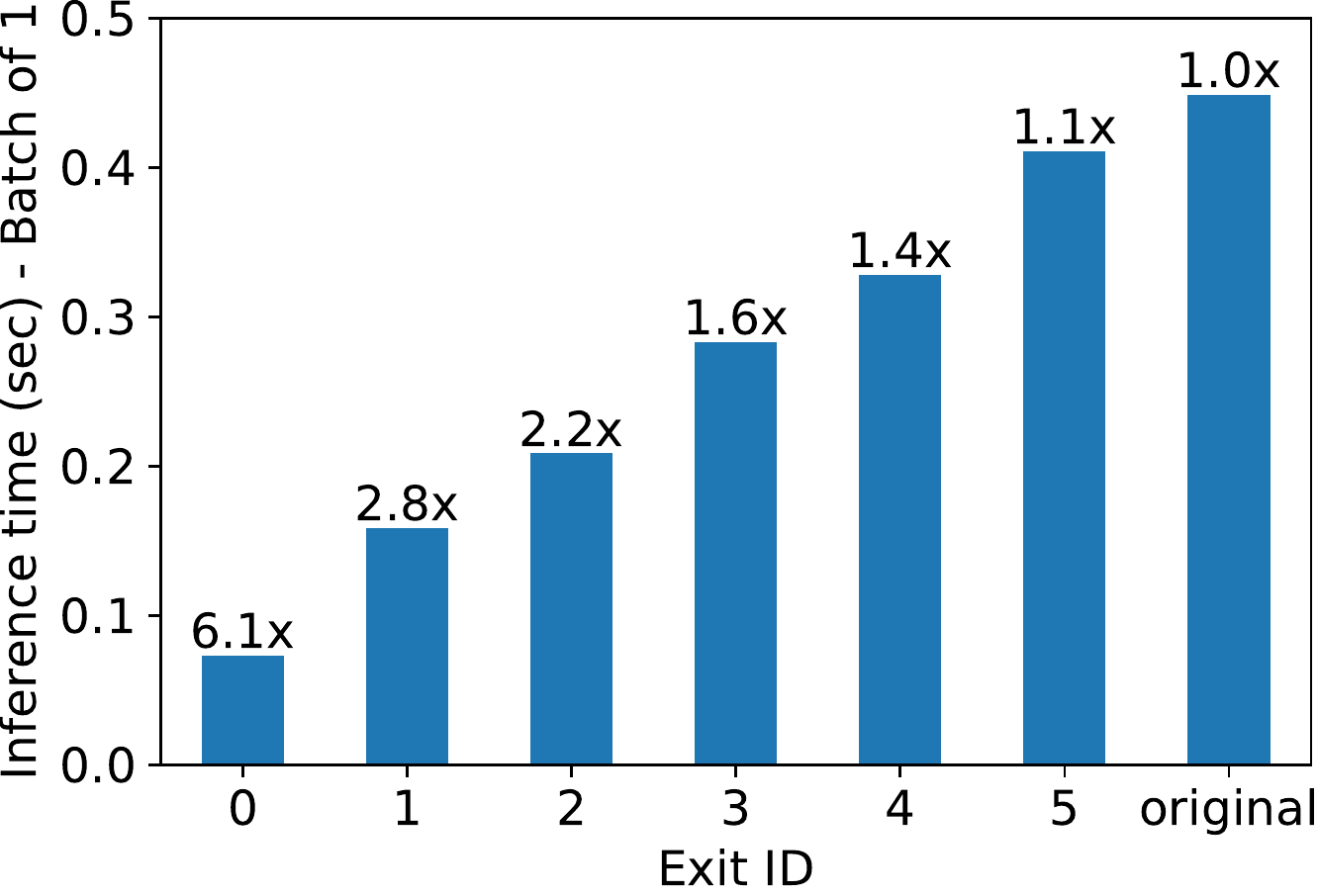}
      \caption{Per-exit inference time.}
      \label{fig:t-a}
    \end{subfigure}%
    \vspace{-0.3cm}
    \caption{Execution times for training and inference of personalised multi-exit ResNet-50 on mobile CPU.}
    \label{fig:times}
    \vspace{-1.5em}
\end{figure}

% In terms of training,
\noindent
\textbf{Training:} On-device personalisation does not require full re-training of the model at hand. With the network's backbone frozen, we only personalise the early classifiers leading to significant gains in performance. As witnessed in  Fig.~\ref{fig:t-b}, multi-exit personalisation can be between $2\times$ and $22.7\times$ faster than full model training, with the main cost allocated to the forward pass. 
This speedup is mainly attributed to not re-training the whole model. This way, we avoid the computational and memory overheads of tracking activations and gradients for the full model.
% , thus making overnight training possible. 

\blue{ \tool aims to train networks overnight, when the device is charged and idle~\cite{sysdesign_fl2019mlsys}.  Our experiments show that such a training on a mobile platform (Jetson's Arm Cortex CPUs) is possible. For instance, to train Exit-2 overnight with 2048 samples only requires 1.7 hours, whereas personalising all the exits simultaneously takes 6.8 hours. Finally, notice that personalisation will not be required frequently, but only when a data distribution shift is detected by the orchestrator.}

% In terms of inference
\noindent
\textbf{Inference:} In Fig.~\ref{fig:t-a}, we observe that the inference time grows linearly with respect to the early-exit FLOPs.
% (Section~\ref{sec:multiexit_model}). 
As a result, on-device personalisation can deliver significant speed-ups, leading to up to $3\times$ higher inference throughput when hard labels are used in our ImageNet-based experiments.

% \subsection{Confidence Threshold}
% \label{sec:conf_thresh_eval}

\begin{figure}[t]
      \centering
    %   \vspace{-0.275cm}
      \vspace{-0.15cm}
      \includegraphics[width=.5\columnwidth]{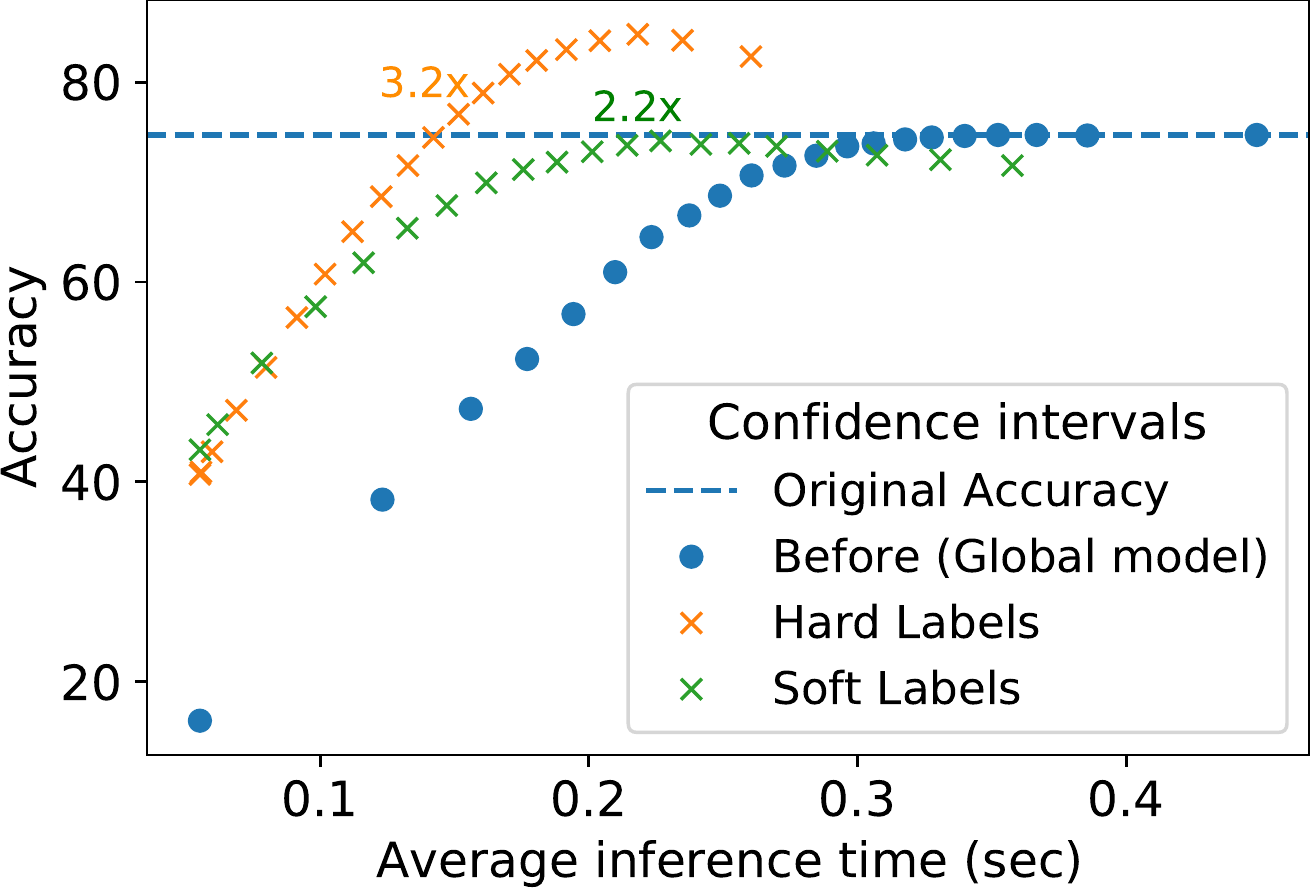}
      \vspace{-0.3cm}
      \caption{Accuracy vs. inference latency for various confidence thresholds.}
      \vspace{-1em}
      \label{fig:confidence}
\end{figure}

\tool allows multiple exits to be attached upon deployment, with the exit of a sample to be chosen at run time by comparing with a confidence threshold. This way, easier examples do not need to pass through the full depth of the CNN, while out-of-distribution inputs (\textit{e.g.}~a new environment) propagate until the last classifier and, therefore, maintaining the original model accuracy. Fig.~\ref{fig:confidence} shows the accuracy-latency trade-off for confidence thresholds {\small $thr_{\text{conf}}\in[0,1]$} (increments of 0.05). We see that \tool's orchestrator can explore this space depending on the device capabilities and performance of the personalised model. For example, when the ground truth is available for personalisation, the orchestrator can choose configurations with similar performance to the original network while achieving $3.2\times$ higher inference throughput ($2.2\times$ with soft labels).

% \begin{itemize}
%     \item Start with experimental setup (server gpu and jetson xavier).
%     \item Accuracy on personalised exits. We can show it as an ablation to see what we get by using hard labels, predictions of last cls as labels and distillation with different temperatures and alpha parameter values. Here the baseline for accuracy should be the backbone exit, as well as the early exits, but without personalisation.
%         \begin{itemize}
%             \item Remind them that we cannot personalise the last classifier as we would have no way of labelling the local dataset then.
%         \end{itemize}
%     \item Latency benefits on Jeton Xavier by early-exiting on confident results. Again the same baseline is needed. We can compare both training and inference times for personalised and non-personalised early-exit networks. In training we don't need to compare to the backbone network as we assume it comes pretrained (at least for Imagenet) and we wouldn't train it on Jetson.
% \end{itemize}

\vspace{-0.3cm}
\section{Discussion and Future Work}
\label{sec:discussion}

%In this work, we have tackled the problem of efficient on-device personalisation with \tool, by using early exits as personalised classifiers for each user.

\blue{
With the wide availability of on-device data and the ever-increasing concern about privacy, on-device training constitutes a strong competitor to centralised solutions. In addition, mobile devices are now equipped with unprecedented compute capabilities.
In this context, ML and systems researchers and developers of today have to rethink the status quo assumptions, \textit{e.g.}~IID-ness of user data, and take advantage of the untapped compute available on users' devices.
}

\blue{
\tool offers a new perspective to the problem of lifelong learning \steve{maybe use continual/lifelong learning as a term here?}\stelios{Done.} through a system that exploits on-device data and multi-exit neural architectures to tailor models to a specific user, improving both accuracy and efficiency. Two of the key components in this endeavour are the ability to adapt CNNs to a specific environment \textit{with or without} labelled data and the automated monitoring of both the training and inference parameters by the system orchestrator, continuously \steve{Too much repetition of continuous}\stelios{Done.} \steve{highlight that monitoring does not impede privacy} exploring opportunities to further improve performance. 
% The system orchestrator automatically monitors both the training and the inference parameters to explore any opportunities to constantly keep improving performance.
}
 
\blue{
At the same time, recent approaches that embrace the heterogeneity of data and the abundance of devices in the wild are becoming increasingly relevant.
Meta-learning~\cite{hospedales2020metalearning} attempts to find good initialisation for personalised models and to leverage expertise on different domains. Federated learning (FL)~\cite{fl2017aistats} aims to train globally accurate models without directly accessing the users' data. 
We believe that our work offers a missing piece in the equation, boosting the efficiency in the deployment of personalised models. \steve{maybe some citations will be helpful on each front.}\stelios{Done.}
}
 
\blue{
As future work, we plan to integrate \tool with the aforementioned approaches to solve emerging challenges, such as avoiding catastrophic forgetting of less frequent knowledge or finding good initialisation for the early classifiers via meta-learning. Incorporation with FL will allow \tool to share the personalised knowledge to constantly improve the global model while respecting user privacy. Moreover, tighter system integration with mobile accelerators will allow \tool to further optimise the energy and latency footprint of AI training and inference. 
}

\vspace{-.5em}
\section{Conclusions}
\label{sec:conclusions}

This paper presents a framework for efficiently personalising CNNs using solely on-device resources. The proposed system introduces multi-exit networks that allow the customisation of the CNN based on the device capabilities and significantly reduce the computational overhead of training. % by updating only a fraction of the network.
Through an efficient on-device training algorithm that leverages the last exit's output to distill personalised knowledge to the earlier exits, the proposed system counteracts the common shortage of ground-truth labels on the user device. Evaluation shows that \tool boosts the accuracy of early exits on user-specific samples while delivering significant speedup for both inference and training, making a decisive step towards on-device personalisation of CNNs on mobile platforms.

\vspace{-0.2cm}

% \balance
\bibliographystyle{ACM-Reference-Format}
\bibliography{references}

\end{document}